%% file: main.tex
\pdfoutput=1

\documentclass[11pt]{article}

\usepackage[]{acl}

\usepackage{times}
\usepackage{latexsym}

\usepackage[T1]{fontenc}

\usepackage[utf8]{inputenc}

\usepackage{microtype}
\usepackage{amsmath}

\usepackage{cleveref}
\usepackage[utf8]{inputenc}

\usepackage{tabularx}
\usepackage{tabu}
\usepackage{booktabs}
\usepackage{multirow}

\usepackage{graphicx} 
\usepackage{float} 
\usepackage{subfigure} 
\usepackage{amssymb}

\usepackage{pifont}
\newcommand{\cmark}{\ding{51}}%
\newcommand{\xmark}{\ding{55}}%
\usepackage{hyperref}
\usepackage{tablefootnote}
\usepackage{xspace}
\iffalse
    \newcommand{\rui}[1]{\textcolor{blue}{}}
\else
    \newcommand{\rui}[1]{\textcolor{blue}{\bf\small [Rui: #1]}}
\fi

\iffalse
    \newcommand{\todo}[1]{\textcolor{blue}{}}
\else
    \newcommand{\todo}[1]{\textcolor{blue}{\bf\small [TODO: #1]}}
\fi

\newcommand{\ours}{\textsc{MultiHiertt}\xspace}
\newcommand{\num}{10,440 }

\title{\ours: Numerical Reasoning over Multi Hierarchical \\Tabular and Textual Data}

\author{Yilun Zhao$^1$ \quad Yunxiang Li$^2$ \quad Chenying Li$^3$ \quad Rui Zhang$^4$ \\
$^1$Yale University \quad $^2$The Chinese University of Hong Kong \\ 
$^3$Northeastern University \quad $^4$Penn State University \\
\texttt{yilun.zhao@yale.edu} \quad \texttt{1155124348@link.cuhk.edu.hk} \\ \texttt{li.chenyin@northeastern.edu} \quad \texttt{rmz5227@psu.edu} \\
}

\begin{document}
\maketitle
\begin{abstract}
\input{main/abstract}
\end{abstract}

\input{main/introduction}

\input{main/related_work}
\input{main/dataset_construction}

\input{main/MT2Net}
\input{main/experiment_result}

\input{main/conclusion}
\input{main/ethics}
\input{main/acknowledge}
\bibliography{anthology,custom}
\bibliographystyle{acl_natbib}

\appendix
\input{appendix/dataset}

\end{document}

%% file: main/abstract.tex
Numerical reasoning over hybrid data containing both textual and tabular content (e.g., financial reports) has recently attracted much attention in the NLP community. However, existing question answering (QA) benchmarks over hybrid data only include a single flat table in each document and thus lack examples of multi-step numerical reasoning across multiple hierarchical tables. To facilitate data analytical progress, we construct a new large-scale benchmark, \ours, with QA pairs over \textbf{Multi} \textbf{Hier}archical \textbf{T}abular and \textbf{T}extual data. \ours is built from a wealth of financial reports and has the following unique characteristics: 1) each document contain multiple tables and longer unstructured texts; 2) most of tables contained are hierarchical; 3) the reasoning process required for each question is more complex and challenging than existing benchmarks; and 4) fine-grained annotations of reasoning processes and supporting facts are provided to reveal complex numerical reasoning. We further introduce a novel QA model termed MT2Net, which first applies facts retrieving to extract relevant supporting facts from both tables and text and then uses a reasoning module to perform symbolic reasoning over retrieved facts. We conduct comprehensive experiments on various baselines. The experimental results show that \ours presents a strong challenge for existing baselines whose results lag far behind the performance of human experts. The dataset and code are publicly available at \url{https://github.com/psunlpgroup/MultiHiertt}.

%% file: main/introduction.tex
\section{Introduction}
\begin{figure}[!t]
    \captionsetup{singlelinecheck= false, justification=justified}
    \includegraphics[width=0.48\textwidth]{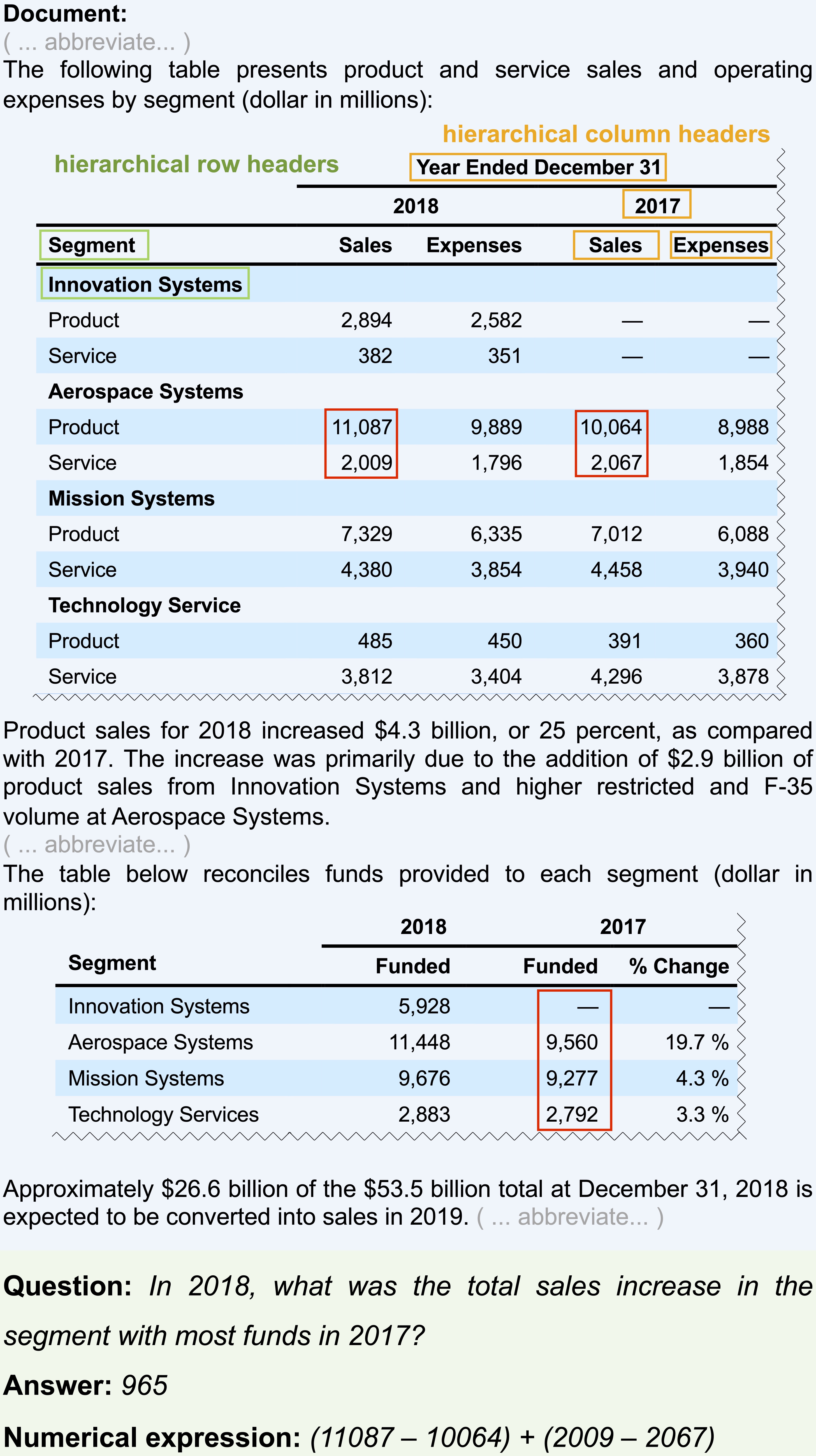}
    \caption{An example of \ours: The system needs to first locate which segment got the most funds in 2017 in the second hierarchical table, then select relevant numbers from the first hierarchical table and generate the correct reasoning program to get the answer. The annotated supporting facts are highlighted in red, and the hierarchical column and row headers are highlighted in orange and green, respectively.}
    \label{example}
\end{figure}

In recent years, as key to many NLP tasks such as QA, there is a flurry of works on numerical reasoning over various types of data including textual data \cite{drop, mathqa, math23k} and tabular data \cite{sciGen, suadaa-etal-2021-towards}. More recently, numerical reasoning over hybrid data containing both textual and tabular content \cite{tatqa, finqa} has attracted much attention. For example, the FinQA dataset \cite{finqa} focuses on questions that require numerical reasoning over financial report pages, e.g., "What portion of the total identifiable net assets is in cash?". Such questions need the system to locate relevant cells in the tabular content and then perform a division operation to get the final answer.

However, existing QA datasets over hybrid data only contain a single flat table in each document~\cite{tatqa, finqa}.
Therefore, they lack examples that require multi-step reasoning processes across multiple paragraphs and hierarchical tables.
Hierarchical tables are widely used in scientific or business documents. A hierarchical table usually contains multi-level headers, which makes cell selection much more challenging because it requires multi-level and bi-dimensional indexing techniques. 
For instance, consider the example of our proposed dataset \ours in Figure~\ref{example}, each table contains both column headers and row headers, which are hierarchical in nature. And ignoring the row / column headers or not reasoning on the entire header hierarchy may lead to the wrong result. For instance, in the given example, if the system simply searched for cells with a flat row header containing "Product" and "Service" and column header containing "2018", it may mistakenly return the value 2,894 and 382 appearing in the beginning of the first table. Additionally, in real life, when analyzing financial reports, professionals such as analysts or investors often refer to multiple hierarchical tables and multiple paragraphs to obtain conclusions. For instance, finding "the segments with most funds in 2017" requires the system to locate and perform numerical reasoning on the second hierarchical table. Then the system should use the results gained from the second table to reason on the first table. However, existing QA datasets lack such examples of reasoning across multiple tables.

To address these shortcomings, we present \ours: an expert-annotated dataset that contains \num QA pairs, along with annotations of reasoning processes and supporting facts. 
To the best of our knowledge, \ours is the first dataset for solving complicated QA tasks over documents containing multiple hierarchical tables and paragraphs.
In addition, to address the challenge of \ours, we propose MT2Net to first retrieve supporting facts from financial reports then generate executable reasoning programs to answer the questions. Our experiments show that MT2Net outperforms all other baselines and achieves 38.43\% F1 score. However, all models still lag far behind the performance of human experts with 87.03\% in F1. It demonstrates \ours presents a strong challenge for existing baseline models and is a valuable benchmark for future research. 

The main contribution of this work can be summarized as follows: 
\begin{itemize}
    \item We propose a new large-scale dataset \ours. It contains \num examples along with fully annotated numerical reasoning processes and supporting facts. A strict quality control procedure is applied to ensure the meaningfulness, diversity, and correctness of each annotated QA example.
    \item Compared with existing datasets, each document in \ours contains multiple hierarchical tables and longer unstructured text. A more complex reasoning process across multiple tables and paragraphs is required to correctly answer the question.
    \item  We propose a novel QA model, MT2Net. The model first applies facts retrieving to extract relevant supporting facts from both hierarchical tables and text. And it then uses a reasoning module to reason over retrieved facts. 
    \item Comprehensive experiments are conducted on various baselines. The experimental results demonstrate that the current QA models still lag far behind the human expert performance, and further research is needed.
\end{itemize}

\begin{table*}[!t]
\small
\centering
\setlength\tabcolsep{4pt}
\begin{tabular}{lrrrrrr}
\toprule
\multicolumn{1}{c}{\multirow{3}{*}{\textbf{QA Dataset}}} & \multicolumn{3}{c}{\multirow{2}{*}{\textbf{Textual \& Tabular Data / Doc 
(DB)}}}                            & \multirow{3}{*}{\textbf{\begin{tabular}[c]{@{}c@{}}Numerical\\ Reasoning\end{tabular}}} & \multirow{3}{*}{\textbf{\# Doc (DB)}} & \multirow{3}{*}{\textbf{\# Questions}} \\
\multicolumn{1}{c}{}                                  & \multicolumn{3}{c}{}                                                                                     &                                                                                          &                                        &                                       \\
\multicolumn{1}{c}{}                                  & \multicolumn{1}{l}{Avg. \# words} & \multicolumn{1}{l}{Table types} & \multicolumn{1}{l}{Avg. \# tables} &                                                                                          &                                        &                                       \\
\midrule
\textbf{Textual QA Dataset} \\
DROP \cite{drop}                                      & 210.0                             & \xmark                          & \xmark                             & \cmark                                                                                   & 6,735                                  & 45,959                                \\
MathQA \cite{mathqa}                                  & 37.9                              & \xmark                          & \xmark                             & \cmark                                                                                   & 37,259                                 & 37,259                                \\
Math23K \cite{math23k}                                & 35.4                              & \xmark                          & \xmark                             & \cmark                                                                                   & 23,161                                 & 23,161                                \\
\\
\textbf{Tabular QA Dataset} \\
WTQ \cite{wtq}                                        & \xmark                            & Flat                            & 1                                  & \xmark                                                                                   &     2,108                                   & 22,033                                      \\
Spider \cite{spider}                                  & \xmark                            & Relational                      & 5.13                               & \xmark                                                                                   & 200                                    & 10,181                                \\
AIT-QA \cite{aitqa}                                   & \xmark                            & Hierarchical                    & 1                                  & \xmark                                                                                   & 116                                    & 515                                   \\
HiTab \cite{hitab}                                    & \xmark                            & Hierarchical                    & 1                                  & few                                                                                   & 3,597                                  & 10,686                                \\
\\
\textbf{Hybrid QA Dataset} \\
HybridQA \cite{hybridqa}         & 2,326.0    & Flat    & 1      & \xmark     & 13,000     &   69,611 \\
MMQA \cite{multimodalqa}                      & 240.7              & Flat                            & 1                                  & \xmark                                                                                      & 29,918                   & 29,918                                \\
GeoTSQA \cite{geotsqa}                                & 52.4                                   & Flat                            & 1.58                               & few                                                                                      & 556                                    & 1,012                                 \\
TAT-QA \cite{tatqa}                                   & 43.6                              & Mostly Flat                          & 1                                  & \cmark                                                                                   & 2,757                                  & 16,552                                \\
FINQA \cite{finqa}                                   & 628.1                            & Flat                            & 1                                  & \cmark                                                                                   & 2,789                                  & 8,281                                 \\
\\
\ours (Ours)  &        1,645.9 & Hierarchical  &   3.89  & \cmark                         &  2,513                      &    \num                   \\ 
\bottomrule
\end{tabular}
\caption{Comparison of \ours with other QA datasets (Doc, DB denote Document and DataBase).}
\label{dataset_comparison}
\end{table*}

%% file: main/related_work.tex
\section{Related Work}
\paragraph{Question Answering Benchmark}
There are numerous QA datasets focusing on text, table/knowledge base (KB), and hybrid data. SQuAD~\cite{rajpurkar2016squad} and CNN/Daily Mail~\cite{hermann2015teaching} are classic datasets for textual data. Table/KB QA datasets mainly focus on structured tables~\cite{wtq, zhong2017seq2sql, spider, nan2021fetaqa} and knowledge bases~\cite{berant2013semantic, yih2015semantic, talmor2018web, xie2022unifiedskg}. And some recent works focus on reasoning over more complex tables including hierarchical tables~\cite{hitab, aitqa}. More recently, there are also some pioneering studies working on QA over hybrid data. Specifically, HybridQA~\cite{hybridqa}, TAT-QA~\cite{tatqa}, and FinQA~\cite{finqa} focus on both textual and tabular data, while MMQA~\cite{multimodalqa} focus on QA over text, tables, and images. In addition, reasoning including numerical reasoning and multi-hop reasoning has gained attention lately. For example, DROP~\cite{drop} is a machine reading comprehension benchmark that requires numerical reasoning on text data. HotpotQA~\cite{yang2018hotpotqa} and HybridQA~\cite{hybridqa} are datasets requiring multi-hop reasoning.

\paragraph{Numerical Reasoning}
Numerical reasoning plays an important role in different NLP tasks~\cite{drop, zhang-etal-2021-noahqa-numerical,finqa,tatqa}. To enhance the model's numerical reasoning ability, some work adapt standard extractive QA models with specialized modules to perform numerical reasoning~\cite{numnet, hu-etal-2019-multi}.  Recent work also focus on probing and injecting numerical reasoning skills to pre-trained language models~\cite{geva2020injecting, lin-etal-2020-birds, zhang-etal-2020-language, berg-kirkpatrick-spokoyny-2020-empirical}. Meanwhile, various benchmarks and models are proposed to solve math word problems~\cite{koncel2016mawps, math23k, mathqa, hendrycksmath2021, hong2021learning, cobbe2021training}. The most recent numerical reasoning QA benchmark over hybrid data are FinQA~\cite{finqa} and TAT-QA~\cite{tatqa}. 


\paragraph{Financial NLP}
Financial NLP has attracted much attention recently. There have been various application in different tasks like risk management~\cite{han-etal-2018-nextgen,theil-etal-2018-word,nourbakhsh2019framework,mai2019deep,wang2019you}, asset management~\cite{filgueiras-etal-2019-complaint,blumenthal-graf-2019-utilizing}, market sentiment analysis~\cite{daudert-etal-2018-leveraging,tabari-etal-2018-causality, buechel-etal-2019-time}, financial event extraction~\cite{ein-dor-etal-2019-financial,zhai-zhang-2019-forecasting} and financial question answering~\cite{lai-etal-2018-simple,maia201818}.
More recently, pre-trained language models are presented for finance text mining~\cite{araci2019finbert,yang2020finbert}. The most relevant work to us is FinQA~\cite{finqa} and TAT-QA~\cite{tatqa}, which both construct a QA dataset acquiring numerical reasoning skills on financial reports with tabular data.

%


%% file: main/dataset_construction.tex
\section{\ours Dataset}
\begin{figure*}[t!]
    \centering
    \includegraphics[width=0.95\textwidth]{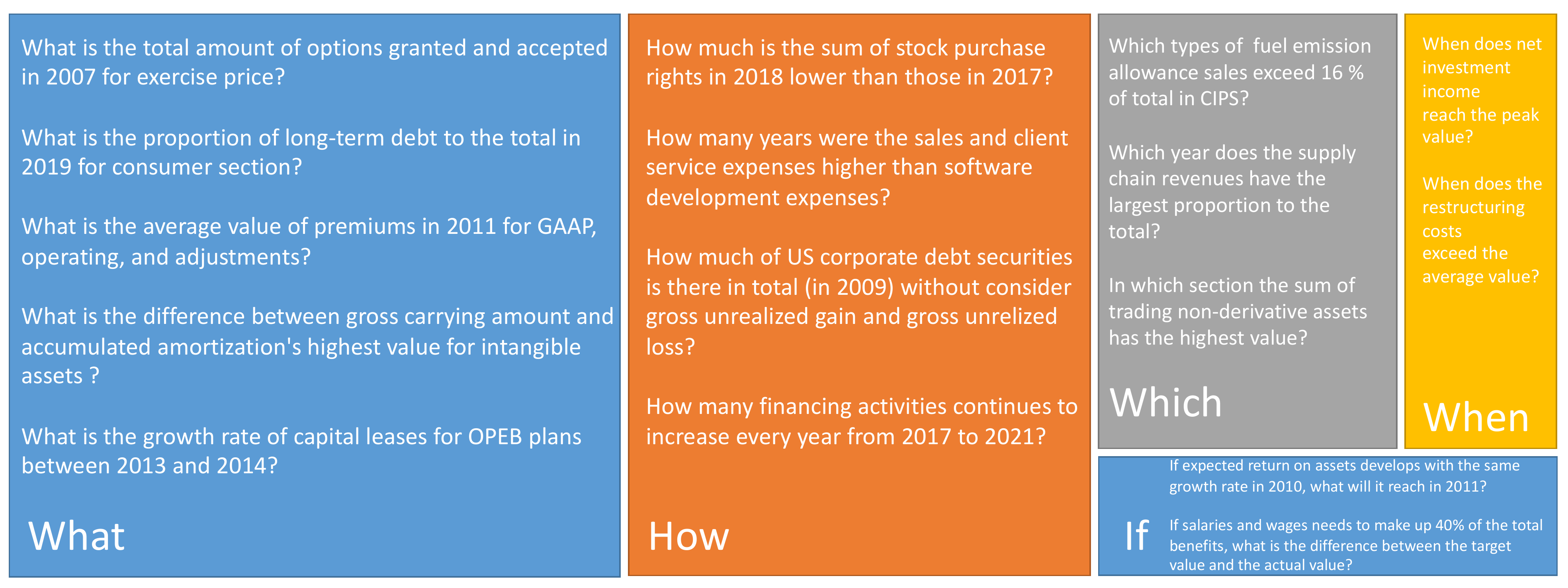}
    \caption{Examples of question by top-$5$ most frequent starting words, where box size represents frequency.}
    \label{question_type}
\end{figure*}

\subsection{Data Collection and Preprocessing} \label{preprocess}
\ours are deployed based on the FinTabNet dataset \cite{fintabnet}, which contains 89,646 pages with table annotations extracted from the annual reports of S\&P 500 companies. For each table contained, the FinTabNet dataset provides a detailed HTML format annotation, in which table hierarchies and cell information such as text and formats can be extracted and post-processed according to HTML tags. 

The raw data is filtered as follows: First, we extract documents with 1 to 4 pages and 2 to 6 tables from FinTabNet. Second, we filter out the documents with limited textual contents. 
Third, as we aim for the numerical reasoning ability, we also exclude documents with tables containing little numerical information. 
Then, we use a pre-processing script to extract the hierarchical structure of each HTML-format table. And we ignore those tables that cannot be handled by the pre-processing script.
As a result, a total of 4,791 documents were selected for further annotation.


\subsection{Question-Answer Pair Annotation}
For each document selected in \S \ref{preprocess}, 
the annotators are required to compose one or two QA examples along with detailed annotation.
The process of annotating each QA example is as follows: 
1) The annotators are first asked to compose a complex question that requires numerical reasoning and is meaningful for helping novices understand the annual reports. The annotators are encouraged to compose questions that require the information from both the textual and tabular content or from multiple tables. 2) For those questions requiring numerical expression, the annotators are then asked to write down the reasoning programs to answer the question. In detail, the annotators are asked to elaborate on the operation steps to answer the question. The definitions of all operations are shown in Table \ref{operation} in Appendix. 3) They are also required to mark all the supporting facts from tabular and textual content for each question.

\subsection{Quality Control}
Strict quality control procedures are designed to ensure the quality of dataset annotation, especially the diversity and meaningfulness of proposed questions. The human evaluation scores and inter-evaluator agreements are reported in Table~\ref{inter-evaluator}. 

\begin{table}[h]
\setlength\tabcolsep{1pt}
\renewcommand{\arraystretch}{1.2}
\small
\begin{center}
\begin{tabular}{lccc}
\toprule
\textbf{Annotation Quality}    & \textbf{\%S $\geq$ 4} & \textbf{Agree} & \multicolumn{1}{c}{\textbf{\begin{tabular}[c]{@{}c@{}}Kappa\\ / 95\% CI\end{tabular}}} \\
\midrule
Question Complexity   & 76.8    & 0.77   &  0.72 / [0.65, 0.79]  \\
Question Correctness  & 93.2    & 0.91   &  0.83 / [0.77, 0.89] \\
Question Meaningfulness  & 91.4  &  0.87  &  0.81 / [0.74, 0.88] \\
Reasoning Correctness &  92.4   & 0.92    & 0.89 / [0.84, 0.94] \\
Support Facts Correctness   & 84.9  & 0.81  & 0.77 / [0.72, 0.82]  \\
Answer Correctness      & 94.0  & 0.93  & 0.90 / [0.87, 0.93]  \\
\bottomrule
\end{tabular}
\end{center}
    \caption{Human evaluation over 100 samples of \ours. Four internal evaluators are asked to rate the samples on a scale of 1 to 5. We report 1) percent of samples that have average score $\geq$ 4 to show high quality of \ours; and 2) percent of agreement and Randolph’s Kappa with 95\% CI \cite{Randolph2005FreeMarginalMK} to show high inter-annotator agreement of \ours.}
    \label{inter-evaluator}
\end{table}

\paragraph{Expert Annotators}
To help improve the annotation process, we first enroll five experts with professional experience in finance. During annotation, they are asked to provide feedback regarding the task instructions and the user experience of the annotation interface, based on which we iteratively modify the annotation guideline and interface design. In the stage of crowd-sourced annotation, we hire 23 graduate students (14 females and 9 males) majoring in finance or similar discipline. Before starting the official annotation process, each annotator is given a two-hour training session to learn the requirements and the annotation interface.

\paragraph{Annotation De-Biasing}
As suggested in previous research~\cite{kaushik, clark, jiang-bansal-2019-avoiding}, consider annotation bias of QA benchmarks is of great significance. During the pilot annotation period, we found that when generating question-answering pairs, annotators may prefer simpler ones. To solve this issue, we use thresholds to restrict the proportions of questions with different numbers of numerical reasoning steps. Meanwhile, the proportions of questions with span selection answer types are set to $\leq$ 20\%. To further increase the diversity of question-answer pair annotation, we also select and include 2,119 QA examples from FinQA~\cite{finqa}. 

\paragraph{Multi-Round Validation}
To further ensure the diversity and correctness of proposed question-reasoning pairs, each document is assigned to three annotators and one verifier in order. For annotators, each is required to first validate the previous annotator's annotation and fix the mistakes if there are. Then, they are asked to create one or two more question-reasoning pairs that are different from the existing ones. After 
each annotator finishes tasks, we assign another verifier with good performance on this project to validate all the annotations.

\subsection{Dataset Analysis}
Core statistics of \ours are reported in Table~\ref{data_details}. Table~\ref{dataset_comparison} shows a comprehensive comparison of related datasets. \ours is the first dataset to study numerical reasoning questions over hybrid data containing multiple hierarchical tables. Compared with TAT-QA and FinQA, documents in \ours contain longer unstructured input text and multiple tables, making the evidence retrieving and reasoning more challenging. And \ours has diverse and complex questions, as illustrated in Figure \ref{question_type}. 

We also analyze supporting facts coverage for each question. In \ours, 1) 10.24\% of the questions only require the information in the paragraphs to answer; 2) 33.09\% of the questions only require the information in one table to answer; 3) 7.93\% require the information in more than one table but without paragraphs to answer; 4) 48.74\% require both the text and table information to answer, and among them, 23.20\% required the information in more than one table. The average number of annotated supporting facts are 7.02. Meanwhile, among those questions with annotated numerical reasoning programs, 28.94\% of them have 1 step; 37.76\% of them have 2 steps; 15.21\% of them have 3 steps; and 18.10\% of them have more than 3 steps. As a result, the average number of numerical reasoning steps is 2.47.

\begin{table}[t!]
\small
\begin{center}
\begin{tabular}{lr}
\toprule
\textbf{Property}                      & \textbf{Value} \\
\midrule
\# Examples (Q\&A pairs with annotation)  &       \num         \\
\# Documents                              &       2,513         \\
Vocabulary                             &       24,193    \\
Avg. \#  Sentences in input text         &      68.06    \\
Avg. \#  Words in input text   &      1,645.9  \\
Avg. \#  Tables per Document     &     3.89           \\
Avg. \#  Rows per Table         &      10.78          \\
Avg. \#  Columns per Table     &       4.97        \\
Avg. \#  Question Length      &      16.78          \\
\midrule
Training Set Size                      &    7,830 (75\%) \\
Development Set Size                   &    1,044 (10\%) \\
Test Set Size                          &    1,566 (15\%) \\
\bottomrule
\end{tabular}
\end{center}
\caption{Core Statistics of \ours.}
\label{data_details}
\end{table}



%% file: main/MT2Net.tex
\section{MT2Net Model} \label{MT2Net}
\begin{figure*}[t!]
    \centering
    \includegraphics[width=0.95\textwidth]{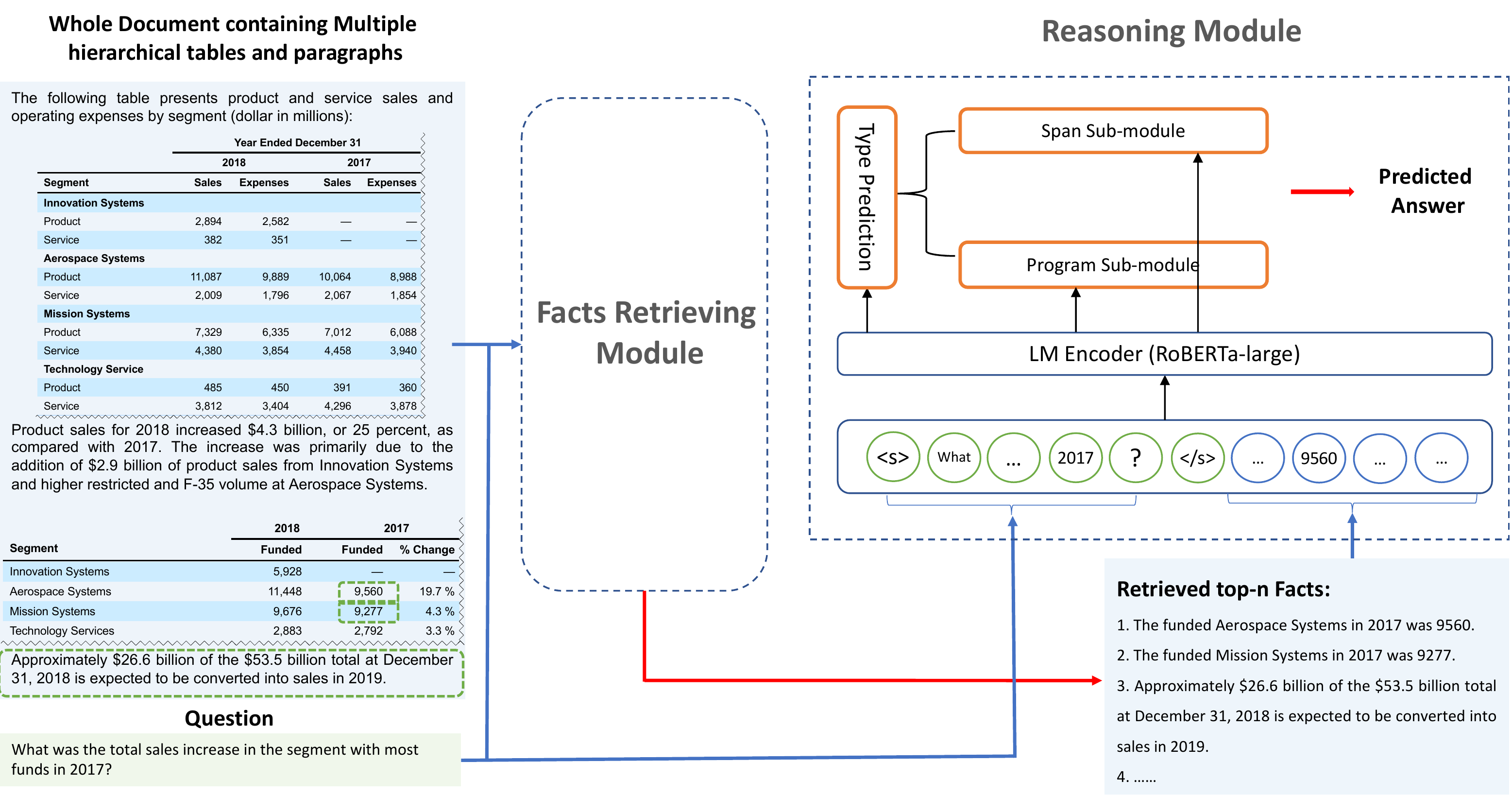}
    \caption{The framework of MT2Net. The model consists of a facts retrieving module and a reasoning module.}
    \label{model_overview}
\end{figure*}

To address the challenge of \ours, we propose a framework named MT2Net. Figure~\ref{model_overview} gives an overview of our proposed model. MT2Net first applies fact retrieving module to extract relevant supporting facts from the hierarchical tables and paragraphs. Then, a reasoning module is adapted to perform reasoning over retrieved facts and get the final answer. 

\paragraph{Fact Retrieving Module}
The whole input text in each document of \ours can exceed 3,000 tokens and contain many numbers, which is beyond the capability of the current popular QA models \cite{bert, roberta}. Therefore, we employ a fact retrieving module to first retrieve the supporting facts from the documents. 
Previous works on hybrid datasets \cite{tatqa, finqa, geotsqa} use templates to flatten each row of the table into sentences. And our facts retrieving module applies similar ideas. However, different from other hybrid datasets, most tables in \ours are hierarchical. Therefore, we turn each cell into a sentence, along with its hierarchical row and column headers. For example, the first data cell in the first table in Figure~\ref{example} is translated as "For Innovation Systems of Segment, sales of product in 2018, Year Ended December 31 is 2,894". 

We concatenate each annotated supporting fact with the question as input to train a BERT-based bi-classifier~\cite{bert}. During the inference stage, the top-$n$ sentences are retrieved as supporting facts. They are reordered according to the order of appearance in the original document. Then they will serve as input to reasoning module.  

\paragraph{Reasoning Module}
We first use pre-trained LMs to encode the retrieved sentences from the facts retrieving module. Then, we divide the answers into two types: arithmetic program and span. For each answer type, we use a unique sub-module to calculate the conditional answer probability $P(\text{answer}|\text{type})$:

\underline{\textit{Program sub-module}}: The structure is similar with the program generator of FinQANet \cite{finqa}. The sub-module aims to generate the executable program to answer the question. Specifically, an LSTM is used for decoding. At each decoding step $T$, the LSTM can generate one token from 1) the numbers from the retrieved, 2) pre-defined operators, and 3) the tokens already generated in the previous steps. After the completion of generation, the sub-module will execute the generated programs and get the predicted answer.

\underline{\textit{Span sub-module}}: The span sub-module aims to select the predicted span candidate, which is a span of retrieved sentences. The answer probability is defined as the product of the probabilities of the start and end positions in the retrieved evidence.

Meanwhile, an extra output layer is used to predict the probability $P(\text{type})$ of each answer type. In particular, we take the output vector $\textsc{[CLS]}$ from LMs as input to compute the probability. In the training stage, the final answer probability is defined as the joint probability over all feasible answer types, i.e., $\sum_{\text{type}} P(\text{type}) \times P(\text{answer}|\text{type})$. Here, both $P(\text{type})$ and $P(\text{answer}|\text{type})$ is learned by the model. In the inference stage, the model first selects the most probable answer type and then uses corresponding sub-modules to predict the answer.

%% file: main/experiment_result.tex
\section{Experiments}

\input{main/baseline_systems}

\subsection{Implementation Details}
For the fact retrieving module, we use BERT-base as the classifier. Since most of the examples in our dataset have less than 7 supporting facts (89.3\%), and we find that longer inputs might lower the performance of the reasoning module, we take the top-$10$ retrieving facts as the retriever results. For the reasoning module, we experiment on using BERT \cite{bert} and RoBERTa \cite{roberta} as the encoder. We use the Adam optimizer \cite{adam} for all models. The training of all models is conducted on RTX 3090s. All the implementation of LMs is based on the huggingface transformers library. To ensure fairness, we set batch size as 32 for all baseline models. 

For Evaluation Metrics, following TAT-QA \cite{tatqa}, we report exact matching (EM) and adopted numeracy-focused F$_1$ \cite{drop}.

\subsection{Human Performance}
To test the performance of the human expert on \ours, we invite another two professionals. We randomly sampled 60 examples from the test set, and ask them to answer the questions individually within three hours.
The results are reported in the last row of Table~\ref{performance}.

\subsection{Model Performance \label{c_performance}}

\begin{table}[t!]
\small
\setlength\tabcolsep{4pt}
\renewcommand{\arraystretch}{1.2}
\begin{tabular}{lrrrr}
\toprule
\multirow{2}{*}{\textbf{}}   & \multicolumn{2}{c}{\textbf{Dev}}                & \multicolumn{2}{c}{\textbf{Test}}               \\
         & \multicolumn{1}{c}{EM} & \multicolumn{1}{c}{F$_1$} & \multicolumn{1}{c}{EM} & \multicolumn{1}{c}{F$_1$} \\
\midrule
Longformer + Reasoning & 2.71                   & 6.93                   & 2.86                   & 6.23                   \\
Facts Retrieving + TAPAS       & 8.94                   & 10.70                  & 7.67                   & 10.04                  \\
Facts Retrieving + NumNet             & 10.32                  & 12.59                  & 10.77                  & 12.02                  \\
TAGOP (RoBERTa-large)            & 19.16                  & 21.08                  & 17.81                  & 19.35                  \\
Facts Retrieving + Seq2Prog          & 26.19                  & 28.74                  & 24.58                  & 26.30                  \\
FinQANet (RoBERTa-large)   & 32.41                  & 35.37                  & 31.72                  & 33.60                  \\          
\midrule
MT2Net (BERT-base)  & 33.68                  & 35.94                  & 32.07     & 33.67             \\
MT2Net (BERT-large)     & 34.03    & 36.13     & 33.25     & 34.98 \\
MT2Net (RoBERTa-base)   & 35.69     & 37.81                  & 34.32                  & 36.17                  \\
MT2Net (RoBERTa-large)    & \textbf{37.05}                  & \textbf{39.96}                  & \textbf{36.22}                  & \textbf{38.43}           \\
\midrule
Human Expert Performance              & \multicolumn{1}{c}{--} & \multicolumn{1}{c}{--} & 83.12                  & 87.03    \\
\bottomrule
\end{tabular}
    \caption{Performance of MT2Net compared with different baseline models on the dev and test sets of \ours. While MT2Net outperforms other baselines, all models perform far behind human experts.
    }
    \label{performance}
\end{table}

Table~\ref{performance} summarizes our evaluation results of different models. We use the same fact retrieving results for all "Retrieving + Reasoning" models. For the fact retrieving module, we have 76.4\% recall for the top-$10$ retrieved facts and 80.8\% recall for the top-$15$ retrieved facts. 

\paragraph{Necessity of applying retrieving-reasoning pipeline}
Directly using an end-to-end pre-trained Longformer model to replace a retrieving module falls far behind. This makes sense because longer input contains much irrelevant numerical information, which makes the reasoning module difficult to learn. 

\paragraph{Necessity of understanding hierarchical table structure}
Both TAGOP and FinQANet perform worse than MT2Net because they ignore the table's hierarchical structure in the retrieving part. Different from ours, which flatten each cell with its header hierarchical structures, both TAGOP and FinQANet flatten each table by rows, losing the table's hierarchical structure information.

\paragraph{Necessity of an effective reasoning module}
Most questions in \ours require models to perform multi-step reasoning and integrate different kinds of operators. 
Generally, the reasoning module generating reasoning programs to get answers performs better than directly generating answers by end-to-end method, i.e. adopted TAPAS. 
Both adopted NumNet and TAGOP perform much worse than MT2Net because they only support limited symbolic reasoning. Specifically, TAGOP can only perform with a single type of pre-defined aggregation operator for each question, and NumNet only supports addition and subtraction operators when performing symbolic reasoning.
By contrast, MT2Net performs better than FinQANet and Seq2Prog because it applies different sub-modules to answer questions with different answer types. 

The results also show that larger pre-trained models have better performance.
This is because they are pre-trained on more financial corpus.
However, all the models perform significantly worse than human experts, indicating \ours is challenging to state-of-the-art QA models and there is a large room for improvements for future research.

\subsection{Further Analysis}
To guide the future directions of model improvement, various performance breakdown experiments on the test set are conducted using the MT2Net (RoBERTa-large) model. Table~\ref{breakdown} shows the results. Generally, the model has a much lower accuracy on questions with more than two numerical reasoning steps. Meanwhile, the model performs poorly on questions requiring cross-table supporting facts.

\begin{table}[t!]
\centering
\small
\begin{tabular}{p{4cm}rr}
\toprule
\textbf{Performance Breakdown}    & \multicolumn{1}{c}{\textbf{EM}}    & \multicolumn{1}{c}{\textbf{F$_1$}}    \\
\midrule
\multicolumn{3}{l}{\textbf{Regarding supporting facts coverage}}                                \\
\\
text-only questions   & 49.26 & 53.29            \\
table-only questions  & 36.77 & 38.55            \\
\quad w/ $\geq$ 2 tables & 24.32  & 24.96          \\
table-text questions  & 33.04  & 35.15                                   \\
\quad w/ $\geq$ 2 tables & 21.04   & 23.36               \\
\midrule
\multicolumn{3}{l}{\textbf{Regarding numerical reasoning steps}}                                          \\
\\
1 step      &     43.62     &   47.80    \\
2 steps     &     34.67     &   37.91    \\
3 steps     &     22.43     &   24.57    \\
> 3 steps   &     15.14      &  17.19                    \\
\midrule
\textbf{Full Results} & \textbf{36.22} & \textbf{38.43} \\
\bottomrule
\end{tabular}
\caption{Results of performance breakdown using MT2Net (RoBERTa-large). The model performance deteriorates as the numbers of tables and reasoning steps increase.}
\label{breakdown}
\end{table}

We further investigate the proposed MT2Net by analyzing error cases. We randomly sample 100 error cases from the results of the MT2Net (RoBERTa-large) model on the test set, and classify them into four main categories as shown in Table~\ref{error_cases}, along with examples. The analysis shows that around 64\% error (Wrong Operand/Span+Missing Operand) is caused by the failure to integrate the supporting facts correctly. Meanwhile, the current model fails to integrate external finance knowledge to answer questions.

\begin{table}[t!]
\small
\begin{tabular}{p{1.3cm}|p{5.5cm}}
\toprule
 &
  {Q: What was the total of premiums granted in the year with the highest GAAP?} \\
 &
  {G: 327 + 415 + 1217} \\
 &
  P: 426 + 517 + 1109 \\
\multirow{-4}{*}{\begin{tabular}[c]{@{}l@{}}Wrong \\ Operand \\ or Span\\ (43\%)\end{tabular}} &
  Explain: Locate the wrong year. \\
  \midrule
 &
  Q:  What was the average value of trading asserts between 2015 and 2018? \\
 &
  G: (1203 + 1437 + 1896 + 1774) / 4 \\
 &
  P: (1203 + 1774) / 2 \\
  
\multirow{-4}{*}{\begin{tabular}[c]{@{}l@{}}Missing \\ Operand \\ (21\%)\end{tabular}}            & Explain: Only account year 2015 and 2018.                  \\
\midrule
 &
  Q: What is the change ratio of corporate debt from 2018 to 2019? \\
 &
  G: (1024 - 979) / 979 \\
\multirow{-3}{*}{\begin{tabular}[c]{@{}l@{}}Wrong \\ Program  \\ (19\%)\end{tabular}} &
  P: 1024 - 979 \\
  \midrule
 &
  Q: What is the earning rate of ATTA stock in 2017? \\
 &
  G: 17.32 / 35.80 \\
 &
  P: 17.32 \\
\multirow{-4}{*}{\begin{tabular}[c]{@{}l@{}}Lack of \\ Domain \\ Knowledge \\ (4\%)\end{tabular}} & Explain: Not know the formula of calculating earning rate. \\
\bottomrule
\end{tabular}
\caption{Examples of error cases and corresponding preparations. Q, G, P denote question, ground truth, and predicted results, respectively.}
    \label{error_cases}
\end{table}

\subsection{Limitations and Future Work}
Although the proposed MT2Net model outperforms other baseline models, it still performs significantly worse than human experts, which reflects the challenge of \ours. Primarily, we find that models do not perform well on certain types of questions: 1) questions requiring reasoning across multiple tables; 2) questions requiring multi-step reasoning; 3) questions requiring reasoning over tables with complex hierarchical structures; and 4) questions requiring external financial knowledge. 

To deal with these challenges, we believe that four main directions of work may be workable: 1) designing a specialized module to handle multi-table reasoning; 2) decomposing a complex question requiring multi-step reasoning into several simpler sub-questions that QA models can handle~\cite{perez-etal-2020-unsupervised, hybridqa}; 3) applying a more advanced table-encoding method. For example, a pre-trained model with specialized table structure-aware mechanisms~\cite{tuta, cheng2021fortap, yang2022tableformer} can be utilized in the facts retrieving module to better understand hierarchical tables; and 4) leveraging structured knowledge \cite{xie2022unifiedskg} to inject external financial knowledge to models.


%% file: main/baseline_systems.tex
\subsection{Baseline Systems}
\paragraph{TAGOP}
TAGOP\footnote{\url{https://github.com/NExTplusplus/tat-qa}} is the baseline model for TAT-QA dataset~\cite{tatqa}. It first uses sequence tagging with the Inside–Outside tagging (IO) approach to extract supporting facts. Then an operator classifier is applied to decide which operator is used to infer the final answer via extracted facts. Different from ours, TAGOP can only perform symbolic reasoning with a single type of pre-defined aggregation operators (e.g. change Ratio, division), and might fail to answer complex questions requiring multi-step reasoning.

\paragraph{FinQANet}
FinQANet\footnote{\url{https://github.com/czyssrs/FinQA}} is the baseline model for FinQA dataset \cite{finqa}. It first uses a BERT-based retriever to take the top-$n$ supporting facts. Then a program generator is applied to generate the reasoning programs to get the final answers. Different from ours, FinQANet ignores the hierarchical structure of tables when linearizing each row of a table. And it is not designed to answer span selection questions.

\paragraph{Longformer + Reasoning module}
To demonstrate the necessity of breaking up models into facts retrieving and reasoning modules, we directly use the pre-trained Longformer-base\footnote{\url{https://github.com/allenai/longformer}} \cite{longformer} as the input encoder in the reasoning module, and encode the whole document.

\paragraph{Fact Retrieving Module + TAPAS}
We employ TAPAS (MASKLM-base)\footnote{\url{https://github.com/google-research/tapas}} \cite{tapas1, tapas2} as a baseline over tabular data. TaPas is pretrained over large-scale tables and associated text from Wikipedia jointly. 
To finetune it, we use the table with most supporting facts along with the answer as input for each example. For the inference stage, the table with most portion of top-15 retrieved facts is used as input.

\paragraph{Fact Retrieving + NumNet}
NumNet+\footnote{\url{https://github.com/llamazing/numnet_plus}} \cite{numnet} has demonstrated its effectiveness on the DROP dataset \cite{drop}. It designs a NumGNN between the encoding and prediction module to perform numerical comparison and numerical reasoning. However, NumNet+ only supports addition and subtraction when performing symbolic reasoning, thus cannot handle those complex questions requiring operators such as division. 

\paragraph{Fact Retrieving Module + Seq2Prog}
A Seq2Prog architecture adopted from baseline of MathQA dataset \cite{mathqa} is used as the reasoning module. 
Specifically, we use a biLSTM encoder and an LSTM decoder with attention.


%% file: main/conclusion.tex
\section{Conclusion}
We have proposed \ours, a new large-scale QA dataset that aims to solve complicated QA tasks that require numerical reasoning over documents containing multiple hierarchical tables and paragraphs. To address the challenge of \ours, we introduce a baseline framework named MT2Net. The framework first retrieves supporting facts from financial reports and then generates executable reasoning programs to answer the question. The results of comprehensive experiments showed that current QA models (best F$_1$: 38.43$\%$) still lag far behind the human expert performance (F$_1$: 87.03$\%$). This motivates further research on developing QA models for such complex hybrid data with multiple hierarchical tables.

%% file: main/ethics.tex
\section{Ethics Considerations}
Data in \ours is collected from the FinQA dataset~\cite{finqa} and FinTabNet dataset~\cite{fintabnet}. FinQA is publicly available under the MIT license\footnote{https://opensource.org/licenses/MIT}. FinTabNet is publicly available under the license CDLA-Permissive-1.0\footnote{https://cdla.dev/permissive-1-0/}. Both licenses permits us to compose, modify, publish, and distribute additional annotations upon the original dataset.

For the internal annotation of \ours, each expert is paid \$20 per hour. For the external  annotation, we hire 23 graduate students majoring in finance or similar disciplines. We regard creating one question-reasoning pair, or validating one document's annotation as a unit task. And we pay around \$1.1 for each unit task. Averagely, an annotator can finish 7 unit tasks per hour after training and practicing. And the hourly rates are in the range of \$6 and \$9 based on the different working speed (above the local average wage of similar jobs). In total, the approximate working hours to annotate \ours dataset is 1500 hours. The whole annotation work lasts about 70 days.

%% file: main/acknowledge.tex
\section*{Acknowledgements}
We appreciate all the annotators' efforts to construct \ours. And we would like to thank the anonymous reviewers and action editors for their constructive discussions and feedback.

%% file: appendix/dataset.tex
\section{Dataset Annotation \label{quality}}
The definitions of all operators used for annotators are shown in Table \ref{operation}.
\begin{table}[h]
\small
\begin{tabular}{lll}
\toprule
Operator & Arguments        & Numerical Expression   \\
\midrule
Add      & number1, number2 & $number1 + number2$    \\
Subtract & number1, number2 & $number1  - number2$   \\
Multiply & number1, number2 & $number1 \times number2$ \\
Divide   & number1, number2 & $number1 \div number2$    \\
Exp      & number1, number2 & $number1^{number2}$   \\
\bottomrule
\end{tabular}
\caption{Definitions of all operations}
\label{operation}
\end{table}

